\documentclass[runningheads]{llncs}

\usepackage{eccv}

\usepackage{xcolor}
\usepackage{hyperref}
\usepackage{eccvabbrv}
\usepackage{microtype}
\usepackage{graphicx}
\usepackage{booktabs}
\usepackage{multirow}
\usepackage{wrapfig}
\usepackage[accsupp]{axessibility}  

\usepackage{hyperref}

\usepackage{orcidlink}

\begin{document}

\title{\textls[-30]{UniGeo: Unifying Geometric Guidance for Camera-Controllable Image Editing via Video Models}}

\titlerunning{Abbreviated paper title}

\author{Hong Jiang\inst{1} \and
Wensong Song\inst{1} \and
Zongxin Yang\inst{2} \and
Ruijie Quan\inst{1} \and
Yi Yang\inst{1}}

\authorrunning{H.~Jiang et al.}

\institute{$^1$ReLER, CCAI, Zhejiang University \quad $^2$DBMI, HMS, Harvard University \\
\vspace{0.15cm}
Project Page: \href{https://mo230761.github.io/UniGeo.github.io/}{\textcolor{magenta}{https://mo230761.github.io/UniGeo.github.io/}}}

\maketitle

\begin{figure}[h]
    \centering
    \includegraphics[width=\textwidth]{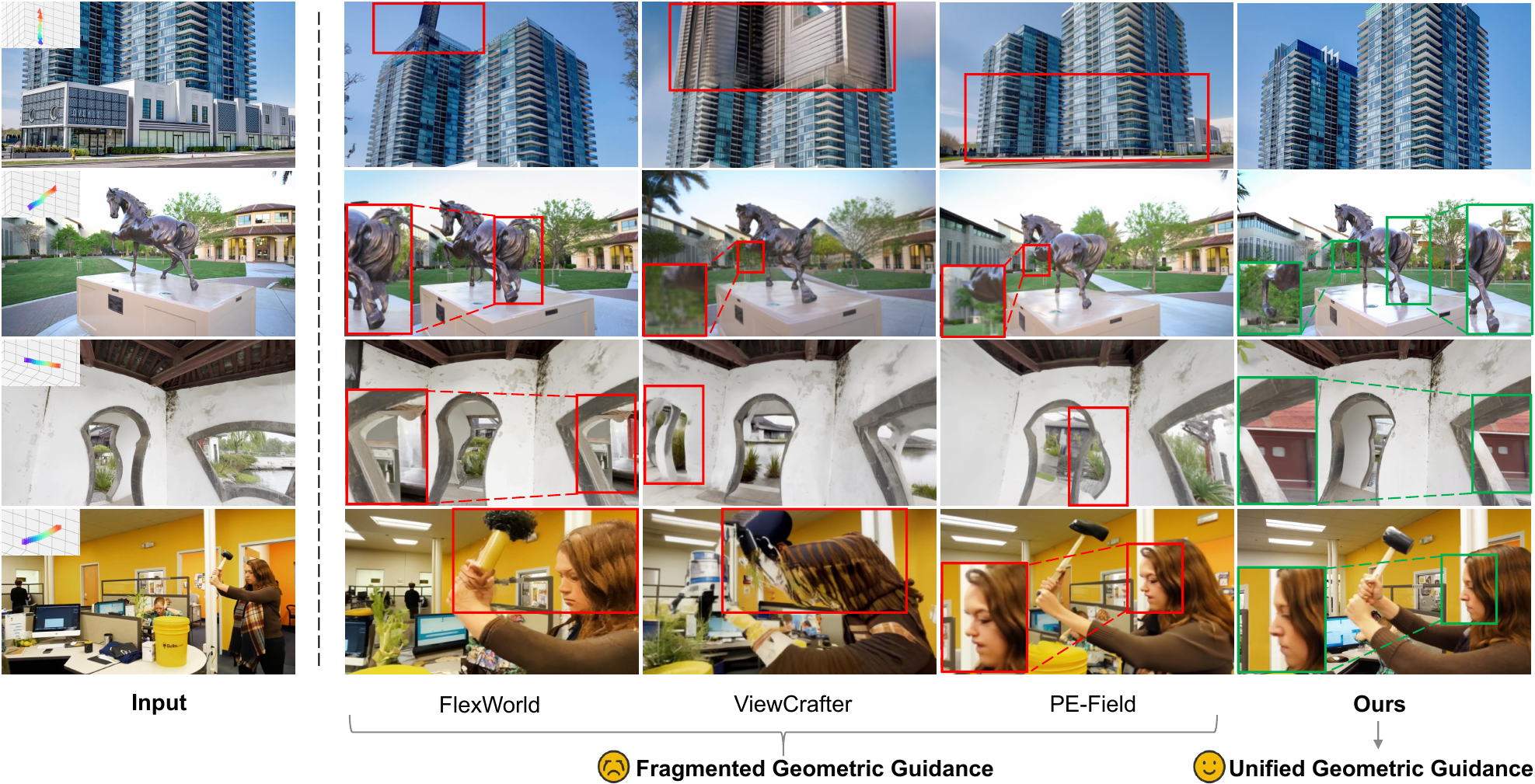}
    \caption{Visual comparisons. Existing methods relying on fragmented geometric guidance often suffer from structural distortions or artifacts under camera motion (highlighted in red). In contrast, by enforcing unified geometric guidance, our \textbf{UniGeo} successfully preserves global scene geometry and structural fidelity (highlighted in green, with selected details enlarged).}
\end{figure}

\begin{abstract}

Camera-controllable image editing aims to synthesize novel views of a given scene under varying camera poses while strictly preserving cross-view geometric consistency. However, existing methods typically rely on fragmented geometric guidance, such as only injecting point clouds at the representation level despite models containing multiple levels, and are mainly based on image diffusion models that operate on discrete view mappings. These two limitations jointly lead to geometric drift and structural degradation under continuous camera motion. We observe that while leveraging video models provides continuous viewpoint priors for camera-controllable image editing, they still struggle to form stable geometric understanding if geometric guidance remains fragmented. To systematically address this, we inject unified geometric guidance across the three levels that jointly determine the generative output: representation, architecture, and loss function. To this end, we propose \textbf{UniGeo}, a novel camera-controllable editing framework. Specifically, at the representation level, UniGeo incorporates a frame-decoupled geometric reference injection mechanism to provide robust cross-view geometry context. Furthermore, at the architecture level, it introduces a geometric anchor attention to align multi-view features, and at the loss function level, it proposes a trajectory-endpoint geometric supervision strategy to explicitly reinforce the structural fidelity of target views. Comprehensive experiments across multiple public benchmarks, encompassing both extensive and limited camera motion settings, demonstrate that UniGeo significantly outperforms existing methods in visual quality and geometric consistency.

  \keywords{Camera-controllable image editing \and Unified geometric guidance \and Video Diffusion Priors}
\end{abstract}

\section{Introduction}
\label{sec:intro}

Camera-controllable image editing \cite{liu2023zero, shi2023zero123++, bai2025positional, chan2023generative, seo2024genwarp, ma2025you} aims to generate transformations of the scene under different viewpoints given changes in camera pose, while maintaining strict cross-view geometric consistency. This capability is critical for applications such as film post-production and robotic perception, among others, directly affecting rendering quality and perception reliability. Unlike general image editing \cite{brack2024ledits++, couairon2022diffedit, meng2021sdedit, mokady2023null, zhang2025icedit, Avrahami_2025_CVPR, kulikov2025flowedit, wang2024taming}, the core challenge here lies not simply in modifying appearance attributes, but in preserving scene structure across views with consistent and seamless generation.

High-quality camera-controllable image editing requires a generative framework that strictly follows camera motion rules while preserving geometric consistency. However, existing methods still face significant challenges: \textbf{(i) Lack of Continuity.} Camera motion is continuous in 3D space, and an ideal framework should reflect the continuous evolution of the scene along the camera trajectory. However, most existing methods \cite{liu2023zero, shi2023zero123++, bai2025positional, seo2024genwarp} are based on image diffusion models and only target mappings between discrete viewpoints, lacking the ability to model continuous camera trajectories. This often results in unstable generation. \textbf{(ii) Lack of Unified Geometric guidance.} Real-world viewpoint changes share consistent geometric correspondences, requiring the model to possess unified guidance over the generation process. However, existing methods \cite{chen2025flexworld, yu2024viewcrafter, guo2024sparsectrl} typically enforce fragmented geometric guidance (such as only injecting point clouds or depth at the representation level). This fragmentation confines geometric guidance to an isolated level, leaving the rest of the model disjointed and unable to form unified correspondences. Consequently, this results in breakages in geometric guidance propagation and ultimately leads to 3D structure collapse.

Based on these observations, we note that video models naturally possess continuous-viewpoint modeling capabilities, providing a potential foundation for camera-controllable image editing \cite{rotstein2025pathways, wu2025chronoedit}. However, even within video models, if geometric guidance is fragmented, the network struggles to form stable geometric understanding across different views \cite{yu2024viewcrafter, chen2025flexworld, he2024cameractrl, wang2024motionctrl}. To systematically address this issue, we draw inspiration from the fundamental levels of model design: representation, architecture, and loss function \cite{Goodfellow-et-al-2016, lecun2015deep, bengio2013representation, domingos2012few}. Since these three levels jointly determine the generation process, this naturally motivates our framework: to ensure global geometric consistency, we systematically inject geometric guidance into each of them.

Motivated by this analysis, we propose \textbf{UniGeo}. Unlike existing methods \cite{yu2024viewcrafter, gao2024cat3d, bai2025positional, chen2025flexworld} that enforce fragmented geometric guidance, UniGeo incorporates unified geometric guidance across representation, architecture, and loss function, systematically rethinking the use of video models for camera-controllable image editing. Specifically, UniGeo achieves this through three tightly coupled modules: \textbf{(i) Frame-Decoupled Point Cloud Injection.} Point clouds encode scene geometry and cross-view correspondences, serving as effective priors. At the representation level, we lift the input image into a trajectory-aligned point cloud sequence and inject it into a video model. Unlike prior methods \cite{zhang2023adding,guo2024sparsectrl,yu2024viewcrafter} that concatenate point clouds along the channel dimension, we decouple them along the frame dimension into independent geometric reference frames. This avoids forcing a strict pixel-to-pixel alignment, preventing the inherently missing points in point clouds from directly corrupting the generated image. \textbf{(ii) Geometric Anchor Attention.} At the architecture level, we introduce an attention mechanism using  first-frame geometric features as \emph{geometry anchors}. Unlike existing I2V methods that only maintain appearance continuity \cite{wan2025, yang2024cogvideox}, our geometric anchors focus on preserving unified geometric consistency across views. During attention interactions, the anchors continuously align features from different viewpoints. \textbf{(iii) Trajectory-Endpoint Geometric Supervision.} At the loss function level, we propose a geometric supervision strategy focusing on camera trajectory endpoints. This shifts the optimization objective from sequence-level consistency to structural fidelity in the target viewpoint. With sparse temporal sampling of key viewpoints, this strategy reduces over-modeling of intermediate frames and applies higher geometric weights to trajectory endpoints, strengthening constraints on target-view 3D structures.

We conduct comprehensive experiments on multiple public video datasets, including RealEstate10K (RE10K) \cite{zhou2018stereo}, Tanks and Temples (Tanks) \cite{knapitsch2017tanks}, DL3DV \cite{ling2024dl3dv}, among others. Unlike previous approaches \cite{yu2024viewcrafter, chen2025flexworld, ren2025gen3c} that split test sets according to video frame intervals, we partition the test videos based on the proportion of newly synthesized regions, categorizing them into \emph{extensive} and \emph{limited camera motion settings}. On RE10K videos with extensive camera motion, the LPIPS decreases from 0.3008 to 0.2377, and on Tanks videos with limited camera motion, the PSNR increases from 16.9580 to 17.8171. These results demonstrate that unified geometric guidance effectively improve cross-view geometric consistency and substantially strengthen camera-controllable image editing across diverse camera motions.

In summary, the main contributions of this work are as follows:
\begin{itemize}
    \item To the best of our knowledge, we propose UniGeo, the first camera-controllable editing framework centered on unified geometric guidance. By overcoming the limitations of relying solely on fragmented geometric guidance and leveraging the continuous-viewpoint prior of video models, our method achieves state-of-the-art (SOTA) performance in camera-controllable image editing under both extensive and limited camera motion settings.
    
    \item To instantiate the unified geometric guidance, we systematically design a tightly coupled model. This comprises, at the representation level, a \textbf{frame-decoupled point cloud injection} mechanism to provide robust geometric context; at the architecture level, a \textbf{geometric anchor attention} module to ensure cross-view structural consistency; and at the loss function level, a \textbf{trajectory-endpoint geometric supervision} strategy to explicitly strengthen the 3D structural fidelity of target viewpoints.
\end{itemize}

\section{Related Work}

\noindent \textbf{Image Editing.} Image editing aims to modify visual content in a controllable manner \cite{huang2025diffusion}. Early diffusion-based approaches rely on training-free inversion \cite{brack2024ledits++, couairon2022diffedit, meng2021sdedit, mokady2023null, parmar2023zero, rout2024semantic, xu2024inversion, ju2023pnp, song2020denoising, hertz2022prompt} and model fine-tuning \cite{bar2022text2live, brooks2023instructpix2pix, huang2024smartedit, kawar2023imagic, yu2025anyedit, zhang2023magicbrush, zhao2024ultraedit}. Recently, this field has been further advanced by large-scale text-to-image foundation models \cite{zhang2025icedit, Avrahami_2025_CVPR, kulikov2025flowedit, wang2024taming, yoon2025splitflow, song2025insert, esser2024scaling, flux2024} and unified auto-regressive architectures \cite{liu2025step1x, wang2025ovisu1, wu2025qwenimagetechnicalreport, chen2025blip3o, cui2025emu35nativemultimodalmodels, deng2025bagel, li2025uniworld, lin2025uniworld, wu2025omnigen2, xiao2025omnigen} for fine-grained semantic control. While excelling at appearance manipulation, these approaches generally lack explicit spatial viewpoint control. Moving beyond general semantic editing, recent works explore camera-controllable image editing \cite{liu2023zero, shi2023zero123++, bai2025positional, gao2024cat3d, wu2024reconfusion, chan2023generative, seo2024genwarp, ma2025you}. However, as these methods are predominantly based on the image diffusion models, they often face challenges in stably modeling continuous camera motion, which may result in geometric inconsistencies across views.

\noindent \textbf{Video Priors for Image Editing.} With the rapid progress of video generation models \cite{blattmann2023stable, brooks2024video, dong2025wan, kong2024hunyuanvideo, ali2025world, opensora2, wang2025wan, zheng2024open}, recent studies have explored directly adapting pretrained video models for image editing tasks. These efforts aim to leverage the continuous temporal priors of video models to maintain structural and semantic consistency during the editing process. Unlike conventional formulations that treat image editing as an independent single-frame generation problem, these approaches directly employ video models to perform image manipulation \cite{rotstein2025pathways, wu2025chronoedit}. However, existing methods generally lack geometric guidance and fail to model the relationship between camera motion and 3D structures, which fundamentally limits their ability to support camera-controllable image editing.

\noindent \textbf{Camera-Controllable Video Generation.} Camera-controllable video generation \cite{bahmani2024vd3d, yang2024direct, zheng2024cami2v, zheng2025vidcraft3, guo2023animatediff, kuang2024collaborative, sun2024dimensionx} incorporates camera motion into video models, enabling synthesized videos to exhibit continuous and consistent viewpoint changes. Existing studies typically introduce conditional signals into the generation process, including encoded camera parameters \cite{wang2024motionctrl, he2024cameractrl, bai2025recammaster, bahmani2025ac3d, van2024generative, xu2024camco}, monocular depth \cite{guo2024sparsectrl, cao2026freeorbit4d}, or view warping \cite{you2024nvs, hou2024training, ren2025gen3c, yu2024viewcrafter, chen2025flexworld}. These signals can be applied to pretrained video models, providing temporal and spatial guidance. However, the geometric guidance in these methods is typically fragmented. They lack unified geometric guidance across the generation pipeline, thus struggle to maintain structural fidelity under continuous camera motion.

\section{Background: Rectified Flow for Video Diffusion Models}

Modern video generation models typically employ a 3D-VAE to compress raw videos into a compact latent space, improving computational efficiency \cite{blattmann2023align, wu2025improved, yang2024cogvideox}. Given an input video $x$, the VAE encoder extracts its latent representation $z_0 = E(x)$, where all subsequent generative modeling is performed. Finally, a decoder reconstructs the generated latents $\hat{z}_0$ back to the pixel space as $\hat{x} = D(\hat{z}_0)$.

Unlike standard diffusion models, recent video foundation models (e.g., Wan \cite{wan2025}) learn the latent distribution via a rectified flow formulation based on flow matching \cite{liu2022flow, albergo2022building, lipman2022flow, esser2024scaling}. For a latent data sample $z_0$ and Gaussian noise $\epsilon \sim \mathcal{N}(0, I)$, an intermediate state $z_t$  at continuous time $t \in [0,1]$ is constructed by linearly blending the data latent and the noise as:
\begin{equation}
z_t = (1 - t) z_0 + t \epsilon ,
\end{equation}

A neural network $F_\theta$ is trained to predict the target velocity field $v = \epsilon - z_0$ that drives this flow. Taking the latent state $z_t$, the timestep $t$, and the conditioning signals (text $y$ and image $c$) as inputs, the training objective is simplified to:
\begin{equation}
\mathcal{L}_\theta =
\mathbb{E}_{t, x, \epsilon}
\left\|
F_\theta([z_t, y, c], t) - (\epsilon - z_0)
\right\|_2^2 .
\end{equation}

\section{UniGeo Model}
\label{sec:blind}

Given an input image $I_0 \in \mathbb{R}^{3 \times H \times W}$ and a camera control prompt, our goal is to synthesize a novel view under the target camera pose, while strictly preserving the underlying 3D geometric structure. 
To this end, we systematically introduce unified geometric guidance at multiple levels to enable accurate camera-controllable image editing.

\begin{figure}[t]
    \centering
    \includegraphics[width=\textwidth]{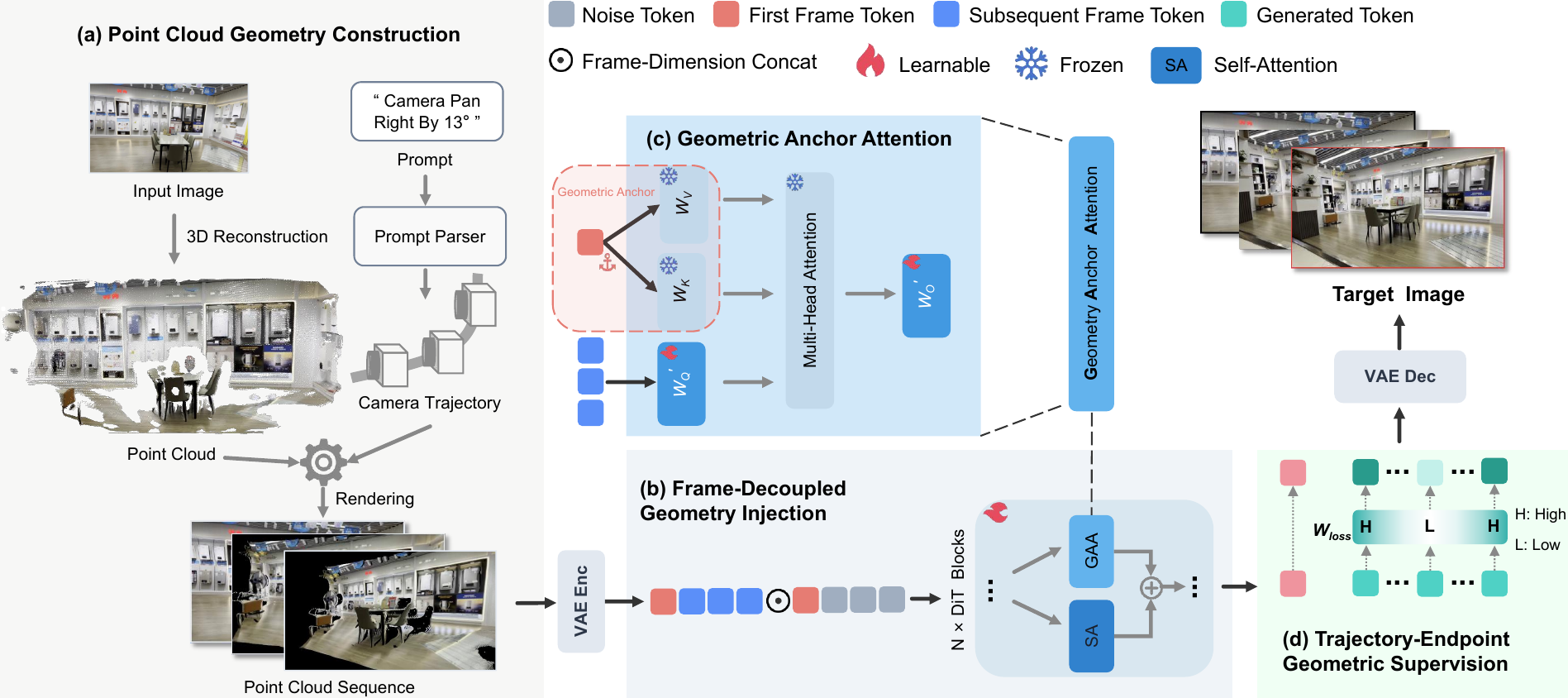}
    \caption{\textbf{UniGeo Framework.} UniGeo incorporates unified geometric guidance through: (a) Geometry Construction: Lifting input images into 3D point cloud sequences. (b) Frame-Decoupled Geometry Injection: Injecting sequences along the frame dimension. (c) Geometric Anchor Attention: Aligning cross-view features using first-frame tokens as anchors. (d) Trajectory-Endpoint Geometric Supervision: Applying higher loss weights to trajectory endpoints versus intermediate frames.}
    \label{fig:pipeline}
\end{figure}

As shown in Fig.~\ref{fig:pipeline}, our approach consists of three modules: Section~\ref{sec:pc_injection} introduces \textbf{Frame-Decoupled Point Cloud Injection}, which constructs point cloud features and injects them into the video model along the frame dimension, including two components: \textit{Point Cloud Geometry Construction} and \textit{Frame-Decoupled Geometry Injection}; Section~\ref{sec:first_frame_attn} presents \textbf{Geometric Anchor Attention}, which continuously aligns geometric features across views; and Section~\ref{sec:tew} describes \textbf{Trajectory-Endpoint Geometric Supervision}, guiding the model to maintain geometric structures at target viewpoints.

\subsection{Frame-Decoupled Point Cloud Injection}
\label{sec:pc_injection}

\noindent \textbf{Point Cloud Geometry Construction.} Directly injecting camera parameters as geometric guidance into diffusion models \cite{bai2025recammaster, wang2024motionctrl, he2024cameractrl} forces the network to implicitly learn the mapping from camera poses to appearance changes. This strategy usually provides only coarse camera controllability and exhibits limited generalization to unseen camera trajectories. Inspired by prior work \cite{yu2024viewcrafter, you2024nvs, hou2024training}, we introduce point cloud sequences as geometric guidance, supplying the video generation model with explicit geometric priors (as shown in Fig.~\ref{fig:pipeline}(a)) .

During training, given an input video $V = [I_0, \ldots, I_{N-1}] \in \mathbb{R}^{N \times 3 \times H \times W}$, where $N$ denotes the number of frames, we first employ VGGT \cite{wang2025vggt} to estimate the camera pose of each frame, yielding the camera trajectory $\mathcal{C} = \{ C_0, \ldots, C_{N-1} \}$. Meanwhile, a point cloud $P_0$ is reconstructed from the first frame $I_0$ using the pre-trained VGGT model \cite{wang2025vggt}.

We then move the virtual camera along the estimated camera trajectory $\mathcal{C} = \{ C_0, \ldots, C_{N-1} \}$ and render the point cloud to obtain a sequence of renderings:
\begin{equation}
R_f = \pi(P_0, C_f), \quad f = 0, \ldots, N-1 ,
\end{equation}
where $C_f$ denotes the camera pose at the $f$-th timestep along the target camera trajectory, and $\pi(\cdot)$ is a differentiable rendering operator. To ensure accurate alignment at the reference view, we explicitly replace the first rendered frame with the original high-fidelity input image, i.e., $R_0 = I_0$.

This procedure produces a rendering sequence that is aligned with the target camera trajectory and serves as geometric guidance for subsequent video generation. Notably, since both the point cloud and the camera poses are estimated by the same model, they naturally reside in a unified coordinate and scale space, which eliminates potential scale inconsistency issues. 

During the inference process, we first convert the user-specified camera control instructions, which indicate the desired camera motion, into the corresponding target camera pose for the ending view.
Then, we interpolate the camera trajectory at uniform intervals between the initial camera pose and the target ending pose to obtain a sequence of camera poses, which is subsequently used to render the point cloud and produce the corresponding rendering sequence.

\noindent \textbf{Frame-Decoupled Geometry Injection.} Motivated by the context conditioning design in DiT-based models \cite{song2025insert, zhang2025icedit, bai2025recammaster}, 
we introduce the rendered point cloud sequence as \emph{frame-decoupled} geometric context and inject it into the video diffusion model by concatenating it with target video tokens (as shown in Fig.~\ref{fig:pipeline}(b)).

Specifically, let $z_t$ denote the latent representation of the target video and $z_s$ denote the latent representation of the rendered point cloud sequence. We apply a patchification operation to obtain $x_t = \mathrm{patchify}(z_t)$ and $x_s = \mathrm{patchify}(z_s)$. These tokens are then concatenated along the frame dimension:
\begin{equation}
x_i = [x_t, x_s]_{\text{frame-dim}} \in \mathbb{R}^{b \times 2f \times s \times d} ,
\end{equation}
which is fed into the DiT backbone as the input token sequence.

This frame-decoupled injection design mitigates the adverse effects of imperfect point cloud priors and allows the geometric context to interact flexibly with target video features throughout the network, naturally supporting our unified guidance and thereby improving cross-view geometric consistency.

\subsection{Geometric Anchor Attention}
\label{sec:first_frame_attn}
To maintain cross-view geometric consistency, we introduce \textbf{Geometric Anchor Attention}, which aligns features across different timesteps using the structural features of the first frame (as shown in Fig.~\ref{fig:pipeline}(c)).

Specifically, given a video sequence of length $N$, we designate the first frame as the geometric anchor and let $X_0$ denote its intermediate feature representation extracted from the backbone network. Its corresponding key $K_0$ and value $V_0$ are obtained via pre-trained projection matrices $W_K$ and $W_V$, i.e., $K_0 = X_0 W_K$ and $V_0 = X_0 W_V$. For any subsequent frame $i \in \{1, \dots, N-1\}$ with features $X_i$, we derive a specific geometric query $(Q_i)' = X_i W'_Q$, where $W'_Q$ is a trainable weight matrix initialized directly from the pre-trained values of the corresponding layer. The Geometric Anchor Attention is then defined as:
\begin{equation}
\mathrm{Attention}((Q_i)', K_0, V_0) = \mathrm{softmax}\Big(\frac{(Q_i)' K_0^\top}{\sqrt{d}}\Big) V_0 .
\end{equation}

The final feature representation is obtained by summing the original self-attention output and the proposed Geometric Anchor Attention:
\begin{equation}
X_i^{\mathrm{out}} = \mathrm{Attention}(Q_i, K_i, V_i) W_O + \alpha \cdot \mathrm{Attention}((Q_i)', K_0, V_0) W'_O ,
\end{equation}
where $Q_i, K_i, V_i$ denote the queries, keys, and values of the original attention mechanism, and $W_O$ is its pre-trained output projection. To ensure training stability and preserve the original generative prior, the new projection matrix $W'_O$ is zero-initialized. Additionally, a scalar weight $\alpha$ is introduced to explicitly control the influence of the geometric guidance.

This design aligns features across timesteps using only two trainable matrices, $W'_Q$ and $W'_O$. It adds minimal computational overhead while serving as the crucial feature-level component of our unified geometric guidance, thereby fundamentally improving cross-view structural consistency.

\subsection{Trajectory-Endpoint Geometric Supervision}
\label{sec:tew}
During temporal modeling, we apply \textbf{sparse temporal sampling} to uniformly select key frames, reducing computation on intermediate frames, and introduce \textbf{Trajectory-Endpoint Geometric Supervision}, which increases loss weights on trajectory-endpoint frames while reducing weights on intermediate frames to enforce geometric stability at the trajectory endpoints (as shown in Fig.~\ref{fig:pipeline}(d)).

Formally, given a sequence of length $N$, we assign a temporally-varying loss weight to each subsequent frame $i$. To explicitly emphasize the trajectory endpoints, the weighting coefficient $w_\text{loss}(i)$ is defined as a quadratic function of the frame's normalized distance to the temporal center:
\begin{equation}
w_\text{loss}(i) = 1 + \gamma \left( \frac{2i}{N-1} - 1 \right)^2, \quad i = 1, \ldots, N-1,
\end{equation}
where $\gamma$ is a hyperparameter that controls the strength of this endpoint penalty.

The final weighted loss for the video sequence is then computed as:
\begin{equation}
\mathcal{L}_\text{weighted} = \sum_{i=1}^{N-1} w_\text{loss}(i) \mathcal{L}_i ,
\end{equation}
where $\mathcal{L}_i$ denotes the original flow matching loss at frame $i$.

Additionally, to further strengthen geometric constraints at the target view, we adopt a \textbf{temporal extension} strategy at the end of the sequence, where the frame corresponding to the target view is duplicated and extended to multiple consecutive timesteps for joint modeling. This design enforces persistent geometric guidance during the final stage of generation, ensuring a stable geometric structure at the target viewpoint.

\begin{table}[t]
    \centering
    \caption{Quantitative comparison of our model with relevant methods under the \textbf{extensive camera motion} setting demonstrates that our model substantially surpasses all relevant baselines across all key metrics. The best and second-best results are demonstrated in \textbf{bold} and \underline{underlined}, respectively.}
    \label{tab:high_change}
    \resizebox{\textwidth}{!}{
        \begin{tabular}{l|cccc|cccc|cccc}
            \toprule
            \multirow{2}{*}{\textbf{Method}} & \multicolumn{4}{c|}{\textbf{DL3DV}} & \multicolumn{4}{c|}{\textbf{RE10K}} & \multicolumn{4}{c}{\textbf{Tanks}} \\
            \cmidrule(lr){2-5} \cmidrule(lr){6-9} \cmidrule(lr){10-13}
             & FID$\downarrow$ & SSIM$\uparrow$ & LPIPS$\downarrow$ & PSNR$\uparrow$ & FID$\downarrow$ & SSIM$\uparrow$ & LPIPS$\downarrow$ & PSNR$\uparrow$ & FID$\downarrow$ & SSIM$\uparrow$ & LPIPS$\downarrow$ & PSNR$\uparrow$ \\
            \midrule
            CameraCtrl \cite{he2024cameractrl} & 172.63 & 0.2703 & 0.4822 & 10.0938 & 124.87 & 0.4320 & 0.4266 & 11.1563 & 116.02 & 0.3787 & 0.3922 & 11.4924 \\
            MotionCtrl \cite{wang2024motionctrl} & 151.00 & 0.3788 & 0.5242 & 10.2664 & 114.65 & 0.5228 & 0.4988 & 10.4614 & 103.97 & 0.4250 & 0.5077 & 10.0663 \\
            ViewCrafter \cite{yu2024viewcrafter} & 146.81 & 0.4700 & 0.4556 & 12.6128 & 97.57 & 0.5905 & 0.3668 & 14.3176 & 105.99 & 0.5418 & 0.4047 & 13.3314 \\
            FlexWorld \cite{chen2025flexworld} & \underline{125.27} & \underline{0.4766} & \underline{0.3726} & \underline{13.3029} & \underline{90.43} & \underline{0.6430} & \underline{0.3008} & \underline{14.3408} & 95.40 & \underline{0.5476} & \underline{0.3395} & \underline{13.8118} \\
            PE-Field \cite{bai2025positional} & 131.93 & 0.4534 & 0.4329 & 12.7060 & 105.34 & 0.6210 & 0.3768 & 13.1684 & \underline{91.96} & 0.5222 & 0.3957 & 13.1063 \\
            \textbf{Ours} & \textbf{113.11} & \textbf{0.4830} & \textbf{0.3248} & \textbf{13.6067} & \textbf{66.67} & \textbf{0.6522} & \textbf{0.2377} & \textbf{14.9723} & \textbf{76.97} & \textbf{0.5574} & \textbf{0.2633} & \textbf{14.4537} \\
            \bottomrule
        \end{tabular}
    }
\end{table}

\section{Experiments}

\subsection{Experimental Settings}

\noindent \textbf{Implementation Details.} We adopt Wan2.2-TI2V-5B \cite{wan2025} as our base video generative model, fine-tuned with a LoRA of rank 256. During training, the frame resolution is fixed at 704$\times$1248, and the video length is set to 29 frames, with the final four frames allocated for persistent modeling of the trajectory endpoints. The model is trained for approximately 10,000 iterations on 4 GPUs, with a learning rate of $1 \times 10^{-4}$ and a total batch size of 4. The hyperparameters $\alpha$ for Geometric Anchor Attention and $\gamma$ for Trajectory-Endpoint Geometric Supervision are set to 1 and 0.01, respectively.

\noindent \textbf{Training Dataset.} To ensure broad scene diversity, we utilize three large-scale datasets for training: DL3DV \cite{ling2024dl3dv}, MannequinChallenge \cite{li2019learning}, and RealEstate10K (RE10K) \cite{zhou2018stereo}. We curated approximately 3,500 samples from DL3DV, 2,500 from MannequinChallenge, and 9,000 from RE10K. Each selected sample consists of 81 frames, from which 29 frames are sparse-temporally sampled for training. All camera trajectories are consistently estimated using the pre-trained VGGT \cite{wang2025vggt}.

\begin{table}[t]
    \centering
    \caption{Quantitative comparison of our model with relevant methods under the \textbf{limited camera motion} setting demonstrates that our model substantially surpasses all relevant baselines across all key metrics. The best and second-best results are demonstrated in \textbf{bold} and \underline{underlined}, respectively.}
    \label{tab:low_change}
    \resizebox{\textwidth}{!}{
        \begin{tabular}{l|cccc|cccc|cccc}
            \toprule
            \multirow{2}{*}{\textbf{Method}} & \multicolumn{4}{c|}{\textbf{DL3DV}} & \multicolumn{4}{c|}{\textbf{RE10K}} & \multicolumn{4}{c}{\textbf{Tanks}} \\
            \cmidrule(lr){2-5} \cmidrule(lr){6-9} \cmidrule(lr){10-13}
             & FID$\downarrow$ & SSIM$\uparrow$ & LPIPS$\downarrow$ & PSNR$\uparrow$ & FID$\downarrow$ & SSIM$\uparrow$ & LPIPS$\downarrow$ & PSNR$\uparrow$ & FID$\downarrow$ & SSIM$\uparrow$ & LPIPS$\downarrow$ & PSNR$\uparrow$ \\
            \midrule
            CameraCtrl \cite{he2024cameractrl} & 141.83 & 0.2403 & 0.4252 & 10.5987 & 122.46 & 0.3994 & 0.3825 & 12.0042 & 97.27 & 0.3697 & 0.3266 & 13.0271 \\
            MotionCtrl \cite{wang2024motionctrl} & 124.60 & 0.3600 & 0.4835 & 10.7530 & 111.43 & 0.5110 & 0.4574 & 11.6774 & 84.60 & 0.4575 & 0.4257 & 12.0949 \\
            ViewCrafter \cite{yu2024viewcrafter} & 103.26 & \underline{0.5205} & 0.3619 & 15.2119 & 84.45 & \underline{0.6370} & 0.2984 & 15.5421 & 73.97 & 0.5930 & 0.3124 & 16.1263 \\
            FlexWorld \cite{chen2025flexworld} & \underline{76.91} & 0.5049 & \underline{0.2827} & \underline{15.4140} & \underline{73.80} & 0.6354 & \underline{0.2573} & \underline{16.1159} & \underline{54.35} & \underline{0.6016} & \underline{0.2418} & \underline{16.9580} \\
            PE-Field \cite{bai2025positional} & 91.83 & 0.4662 & 0.3508 & 14.2761 & 81.76 & 0.6188 & 0.3189 & 15.4383 & 65.49 & 0.5681 & 0.3003 & 16.0821 \\
            \textbf{Ours} & \textbf{69.05} & \textbf{0.5271} & \textbf{0.2065} & \textbf{16.3740} & \textbf{51.73} & \textbf{0.6650} & \textbf{0.1730} & \textbf{17.2989} & \textbf{40.55} & \textbf{0.6278} & \textbf{0.1526} & \textbf{17.8171} \\
            \bottomrule
        \end{tabular}
    }
\end{table}

\noindent \textbf{Testing and Evaluation.} We evaluate our method on the test sets of RE10K, Tanks and Temples (Tanks) \cite{knapitsch2017tanks}, DL3DV, and MannequinChallenge. For RE10K, Tanks, and DL3DV, we categorize camera motion based on the proportion of newly synthesized regions (mask area) in the final frame of the point cloud rendering: videos with a mask ratio $>$ 35\% are classified as \emph{extensive camera motion}, while the remainder are deemed \emph{limited camera motion}. This 35\% threshold is empirically chosen to distinguish significant viewpoint changes. We randomly select 50 video samples from each category to ensure a balanced and representative evaluation. For MannequinChallenge, we randomly select 50 samples due to the inherent complexity of human-centric scenes. We adopt PSNR, SSIM \cite{wang2004image}, LPIPS \cite{zhang2018unreasonable}, and FID \cite{heusel2017gans} as primary evaluation metrics.

\begin{figure}[t]
    \centering
    \includegraphics[width=\textwidth]{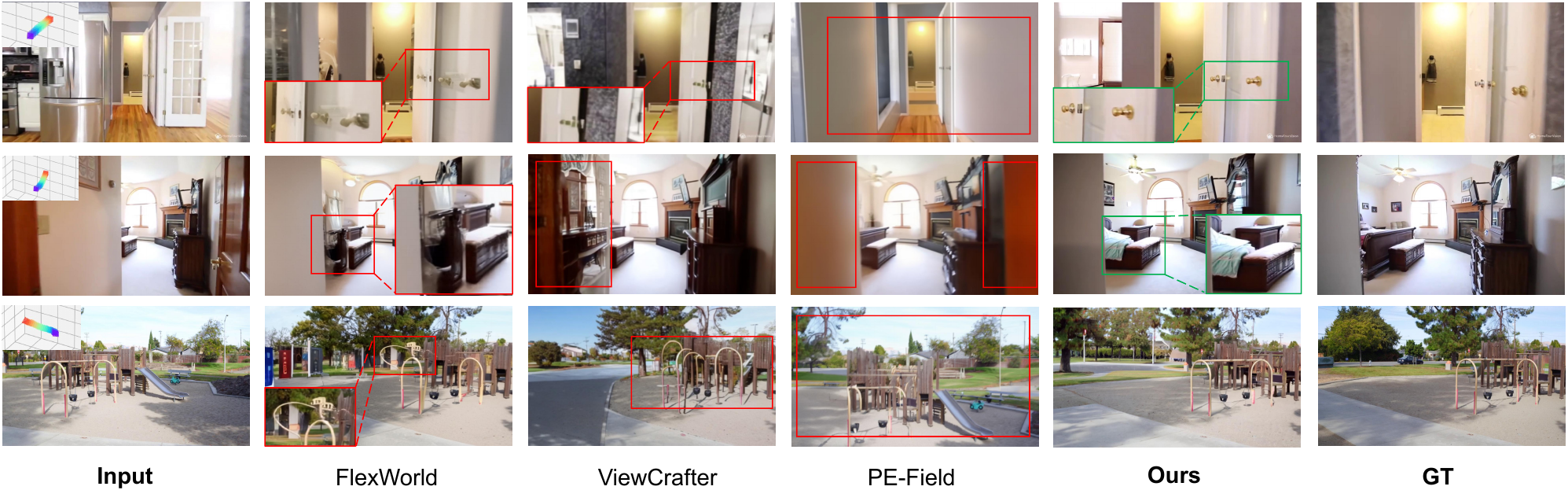}
    \caption{Qualitative comparison under the \textbf{extensive camera motion} setting. 
Compared with other methods, our approach better preserves the geometric structure of the scene under extensive camera motion, effectively avoiding structural duplication.}
    \label{fig:high}
\end{figure}

\subsection{Comparisons with relevant methods}

\noindent \textbf{Quantitative comparisons.}
We evaluate our method against CameraCtrl \cite{he2024cameractrl}, MotionCtrl \cite{wang2024motionctrl}, ViewCrafter \cite{yu2024viewcrafter}, FlexWorld \cite{chen2025flexworld}, and PE-Field \cite{bai2025positional} on DL3DV, RE10K, and Tanks under both extensive and limited camera motion settings, as shown in Tab.~\ref{tab:high_change} and Tab.~\ref{tab:low_change}. Our method achieves the best performance across all key metrics, demonstrating strong capability to maintain high fidelity and perceptual quality under varying levels of camera motion.  In addition, on the MannequinChallenge dataset \cite{li2019learning} (Tab.~\ref{tab:mannequin}), our method also attains the best results. These results demonstrate that our method effectively preserves cross-view geometric consistency and substantially improves generation quality.

\begin{figure}[t]
    \centering
    \includegraphics[width=\textwidth]{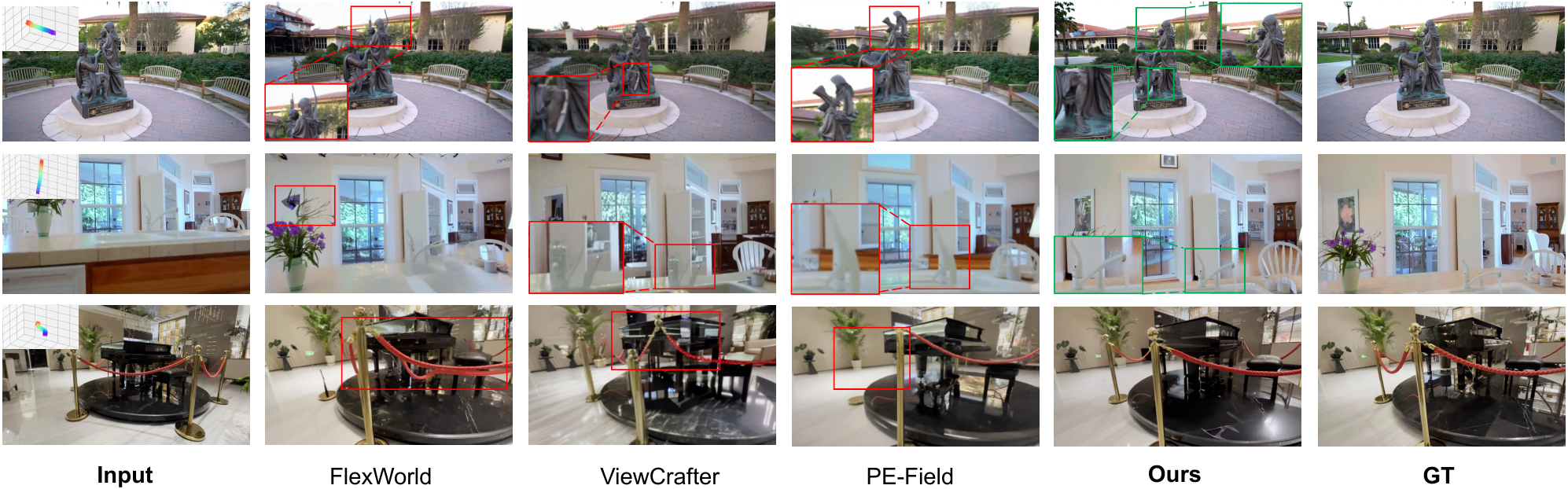}
    \caption{Qualitative comparison under the \textbf{limited camera motion} setting. 
Our method maintains stable spatial layouts and scene structural consistency across views, while better preserving fine-grained scene details.}
    \label{fig:low}
\end{figure}

\begin{wraptable}[8]{r}{0.55\columnwidth}
    \centering
    \vspace{-26pt}
    \setlength{\tabcolsep}{1.2pt} 
    \caption{Results on \textbf{MannequinChallenge}.}
    \label{tab:mannequin}
    {\scriptsize
    \begin{tabular}{l|cccc}
        \toprule
        \textbf{Method} & FID$\downarrow$ & SSIM$\uparrow$ & LPIPS$\downarrow$ & PSNR$\uparrow$ \\
        \midrule
            CameraCtrl \cite{he2024cameractrl} & 207.43 & 0.3031 & 0.5316 & 9.7386 \\
            MotionCtrl \cite{wang2024motionctrl} & 200.98 & 0.4258 & 0.5596 & 9.6773 \\
            ViewCrafter \cite{yu2024viewcrafter} & 202.93 & 0.4675 & 0.4830 & 11.0729\\
            FlexWorld \cite{chen2025flexworld} & \underline{183.69} & \underline{0.5434} & \underline{0.4111} & \underline{12.7678}\\
            PE-Field \cite{bai2025positional} & 185.76 & 0.5301 & 0.4634 & 12.2003  \\
            \textbf{Ours} & \textbf{172.63} & \textbf{0.5546} & \textbf{0.3735} & \textbf{12.7902} \\
        \bottomrule
    \end{tabular}
    \vspace{-22pt}
    }
\end{wraptable}
\noindent\textbf{Qualitative comparisons.}
Fig.~\ref{fig:high} and Fig.~\ref{fig:low} present qualitative comparisons under the extensive and limited camera motion settings, respectively. It can be observed that, under camera motion, existing methods often struggle to preserve the geometric structure of the scene during novel view generation, leading to artifacts such as duplicated structures, distorted geometric relationships, and locally incoherent content, especially when the camera motion becomes extensive. In contrast, our method better maintains geometric consistency across views and produces more natural and coherent novel-view results, effectively alleviating structural degradation and visual inconsistencies.

\begin{figure}[t]
    \centering
    \includegraphics[width=\textwidth]{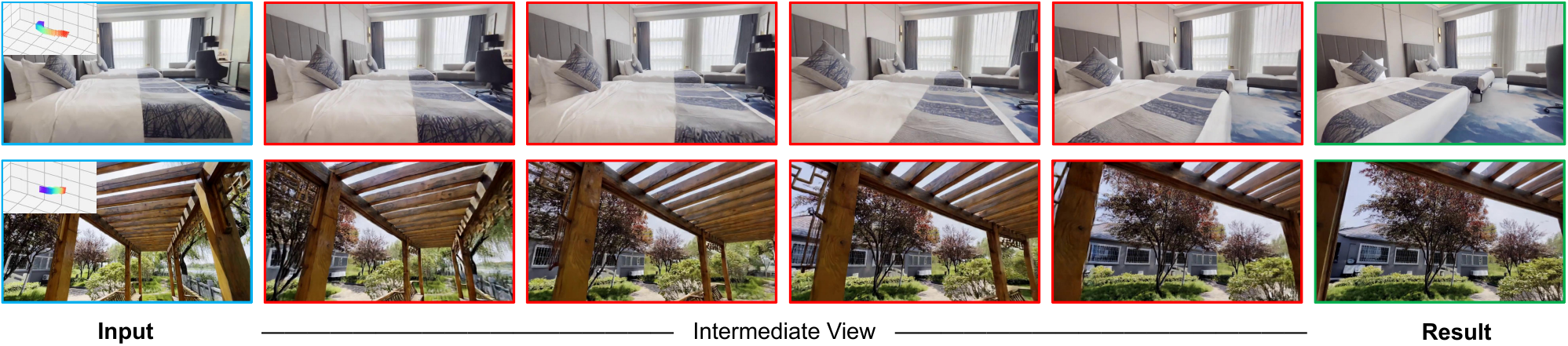}
    \caption{Our approach models continuous camera motion characteristics. Sequences are shown from left to right: the input image (blue), intermediate frames reflecting the trajectory (red), and the final novel view (green).}
    \label{fig:intermediate_process}
\end{figure}

\begin{figure}[t]
    \centering
    \includegraphics[width=\textwidth]{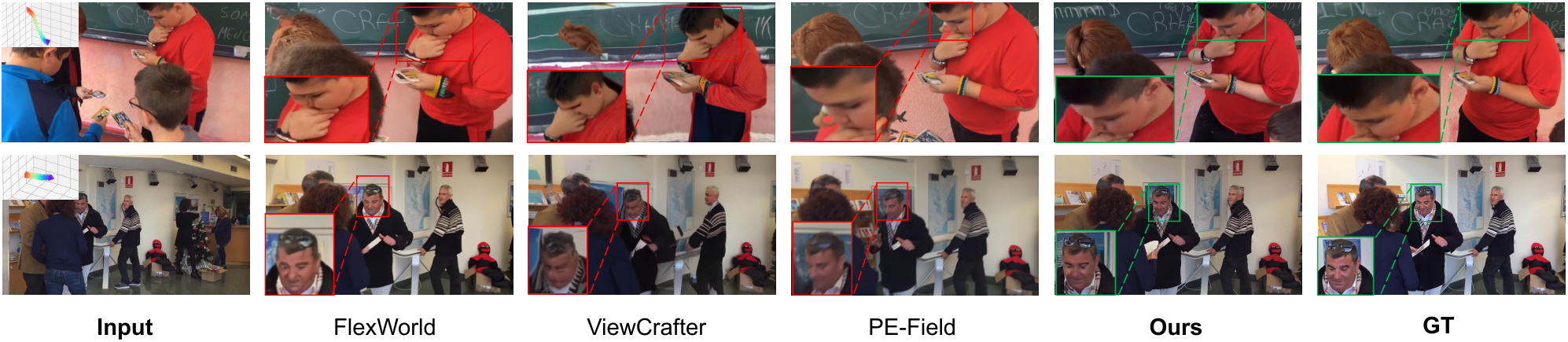}
    \caption{Qualitative comparison on the \textbf{MannequinChallenge} dataset. 
Under camera motion, our method achieves more stable identity preservation compared with other methods, maintaining more consistent appearance.}
    \label{fig:person}
\end{figure}

In addition, Fig.~\ref{fig:person} shows qualitative results on the MannequinChallenge \cite{li2019learning} dataset, which mainly focuses on human-centric scenes. Compared with other methods, our approach exhibits more stable identity preservation across views and thus produces more reliable results under camera motion.

\noindent\textbf{Intermediate Trajectory Visualization.}
Furthermore, to provide deeper insights into our generation process, we visualize the intermediate synthesized frames along the camera trajectory in Fig.~\ref{fig:intermediate_process}. This visualization demonstrates how our model smoothly and accurately models the continuous geometric transformations dictated by the camera motion. By maintaining structural coherence throughout the intermediate process, our approach aligns with the camera motion characteristics, ensuring precision in the final rendered views.

\subsection{Ablation Study}

To verify the effectiveness of the proposed designs, we conduct comprehensive ablation studies on three key components: Frame-Decoupled Point Cloud Injection (FDPCI), Geometric Anchor Attention (GAA), and Trajectory-Endpoint Geometric Supervision (TEGS). All experiments are conducted on the DL3DV \cite{ling2024dl3dv} dataset under extensive and limited camera motion settings. Due to space constraints, additional ablation studies are provided in the supplementary material.

\begin{table}[t]
    \centering
    \caption{Ablation study on the \textbf{DL3DV dataset} under extensive and limited camera motion settings. We evaluate Frame-Decoupled Point Cloud Injection (FDPCI), Geometric Anchor Attention (GAA) and Trajectory-Endpoint Geometric Supervision (TEGS). The best results are highlighted in \textbf{bold}.}
    \label{tab:ablation_comp}
    {\scriptsize  
    \begin{tabular}{l|cccc|cccc}
        \toprule
        \multirow{2}{*}{\textbf{Setting}} 
        & \multicolumn{4}{c|}{\textbf{Extensive Camera Motion}} 
        & \multicolumn{4}{c}{\textbf{Limited Camera Motion}} \\
        \cmidrule(lr){2-5} \cmidrule(lr){6-9}
        & FID$\downarrow$ & SSIM$\uparrow$ & LPIPS$\downarrow$ & PSNR$\uparrow$ 
        & FID$\downarrow$ & SSIM$\uparrow$ & LPIPS$\downarrow$ & PSNR$\uparrow$ \\
        \midrule
        w/o FDPCI & 121.52 & 0.4270 & 0.3491 & 12.9989 & 75.79 & 0.4941 & 0.2218 & 15.7771 \\
        w/o GAA & 115.83 & 0.4683 & 0.3284 & 13.3682 & 70.44 & 0.5060 & 0.2088 & 15.9162 \\
        w/o TEGS & 119.43 & 0.4588 & 0.3379 & 13.0054 & 69.15 & 0.5050 & 0.2122 & 15.7987 \\
        Ours & \textbf{113.11} & \textbf{0.4830} & \textbf{0.3248} & \textbf{13.6067} & \textbf{69.05} & \textbf{0.5271} & \textbf{0.2065} & \textbf{16.3740} \\
        
        \bottomrule
    \end{tabular}
    }
\end{table}

\begin{figure}[t]
    \centering
    \includegraphics[width=\textwidth]{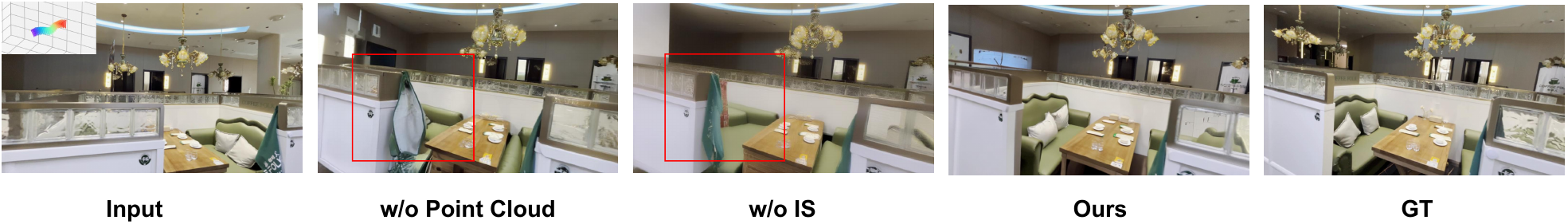}
    \caption{Qualitative results of the ablation study.
Without point cloud or intermediate supervision, the generated results suffer from object duplication, incorrect placement, and increased blur, leading to degraded geometric consistency.}
    \label{fig:ab}
\end{figure}

\noindent \textbf{Frame-Decoupled Point Cloud Injection.} 
We first evaluate the effect of injecting point cloud sequences aligned with the target camera trajectory along the frame dimension as geometric priors. As shown in Table~\ref{tab:ablation_comp}, removing FDPCI leads to a significant performance drop (e.g., LPIPS increases by 0.02 on average and SSIM drops by 0.06 in extensive motions), indicating its crucial role in maintaining structural consistency and perceptual quality. Qualitatively, as shown in Fig.~\ref{fig:ab}, this approach effectively avoids object duplication and positional errors, maintaining more stable geometry under viewpoint changes.

\noindent \textbf{Geometric Anchor Attention.} 
We then evaluate the effect of Geometric Anchor Attention. As demonstrated in Table~\ref{tab:ablation_comp}, the removal of GAA results in noticeable degradation across all metrics, proving that introducing the first-frame geometric features as anchors explicitly aligns cross-view features and preserves geometric structure. To further explore its optimal setting, we conduct a hyperparameter analysis on the attention weight $\alpha$, detailed in Table~\ref{tab:ablation_alpha}. Consistently applying the anchor attention improves generation quality, achieving the best performance at a balanced weight of $\alpha=1.0$. Setting $\alpha$ too low ($\alpha=0.1$) weakens the geometric alignment, while setting it too high ($\alpha=1.5$) overly constrains the features, leading to slight performance drops.

\begin{table}[t]
    \centering
    \caption{Hyperparameter analysis on the weight $\alpha$ of Geometric Anchor Attention (GAA) under extensive and limited camera motion settings on the \textbf{DL3DV dataset}. The best results are highlighted in \textbf{bold}.}
    \label{tab:ablation_alpha}
    \resizebox{\textwidth}{!}{  
    \begin{tabular}{l|cccc|cccc}
        \toprule
        \multirow{2}{*}{\textbf{Setting}} 
        & \multicolumn{4}{c|}{\textbf{Extensive Camera Motion}} 
        & \multicolumn{4}{c}{\textbf{Limited Camera Motion}} \\
        \cmidrule(lr){2-5} \cmidrule(lr){6-9}
        & FID$\downarrow$ & SSIM$\uparrow$ & LPIPS$\downarrow$ & PSNR$\uparrow$ 
        & FID$\downarrow$ & SSIM$\uparrow$ & LPIPS$\downarrow$ & PSNR$\uparrow$ \\
        \midrule
        FDPCI + TEGS + GAA ($\alpha=0.1$) & 115.17 & 0.4476 & 0.3343 & 13.2395 & 70.02 & 0.5096 & 0.2115 & 16.2663 \\
        FDPCI + TEGS + GAA ($\alpha=0.5$) & 113.42 & 0.4636 & 0.3311 & 13.3267 & 69.48 & 0.5084 & 0.2066 & 15.9713 \\
        FDPCI + TEGS + GAA ($\alpha=1.0$) (ours) & \textbf{113.11} & \textbf{0.4830} & \textbf{0.3248} & \textbf{13.6067} & \textbf{69.05} & \textbf{0.5271} & \textbf{0.2065} & \textbf{16.3740} \\
        FDPCI + TEGS + GAA ($\alpha=1.5$) & 114.80 & 0.4718 & 0.3252 & 13.6044 & 70.88 & 0.5103 & 0.2108 & 16.0774 \\
        \bottomrule
    \end{tabular}
    }
\end{table}

\noindent \textbf{Trajectory-Endpoint Geometric Supervision.} Finally, we investigate the design of the Trajectory-Endpoint Geometric Supervision (TEGS). Incorporating TEGS significantly improves structural fidelity at target views and enhances geometric consistency compared to the baseline without it (w/o TEGS in Table~\ref{tab:ablation_comp}). To determine the optimal supervision intensity, we further evaluate different hyperparameter settings for TEGS in Table~\ref{tab:ablation_tegs}. The results show that the best geometric consistency and visual quality are achieved under our configuration.

\begin{table}[t]
    \centering
    \caption{Hyperparameter analysis on the weight $\gamma$ of Trajectory-Endpoint Geometric Supervision (TEGS) under extensive and limited camera motion settings on the \textbf{DL3DV dataset}. The best results are highlighted in \textbf{bold}.}
    \label{tab:ablation_tegs}
    \resizebox{\textwidth}{!}{  
    \begin{tabular}{l|cccc|cccc}
        \toprule
        \multirow{2}{*}{\textbf{Setting}} 
        & \multicolumn{4}{c|}{\textbf{Extensive Camera Motion}} 
        & \multicolumn{4}{c}{\textbf{Limited Camera Motion}} \\
        \cmidrule(lr){2-5} \cmidrule(lr){6-9}
        & FID$\downarrow$ & SSIM$\uparrow$ & LPIPS$\downarrow$ & PSNR$\uparrow$ 
        & FID$\downarrow$ & SSIM$\uparrow$ & LPIPS$\downarrow$ & PSNR$\uparrow$ \\
        \midrule
        FDPCI + GAA + TEGS ($\gamma=0.001$) & 119.58 & 0.4394 & 0.3484 & 13.0486 & 70.98 & 0.4883 & 0.2327 & 15.5676 \\
        FDPCI + GAA + TEGS ($\gamma=0.01$) (ours) & \textbf{113.11} & \textbf{0.4830} & \textbf{0.3248} & \textbf{13.6067} & \textbf{69.05} & \textbf{0.5271} & \textbf{0.2065} & \textbf{16.3740} \\
        FDPCI + GAA + TEGS ($\gamma=0.1$) & 115.60 & 0.4616 & 0.3331 & 13.2588 & 71.57 & 0.4993 & 0.2169 & 15.8824 \\
        \bottomrule
    \end{tabular}
    }
\end{table}

\begin{figure}[!t]
    \centering
    \includegraphics[width=\textwidth]{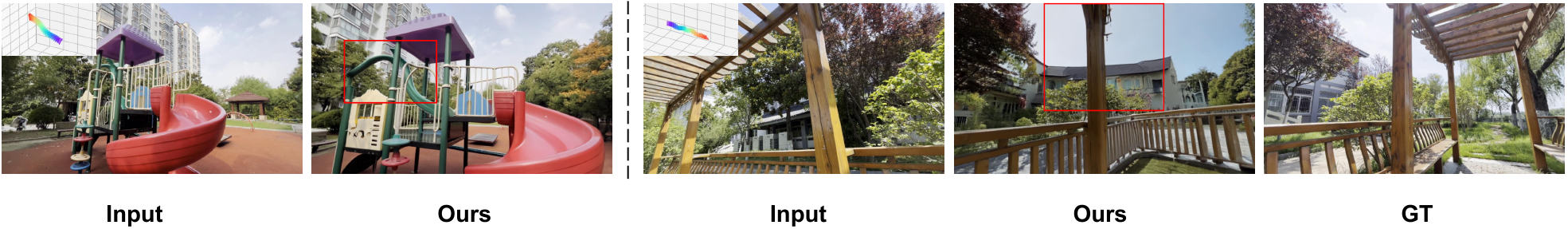}
    \caption{Failure cases: Left—complex objects challenge geometry and texture preservation; Right—extreme camera changes impede geometric consistency.}
    \label{fig:limition}
\end{figure}

Additionally, we conduct a qualitative control experiment where geometric supervision is applied \textit{only} at trajectory endpoints, leaving intermediate frames completely unconstrained (denoted as ``w/o intermediate supervision (w/o IS)'' in Fig.~\ref{fig:ab}). As demonstrated, this extreme setting produces substantially blurrier results, indicating that completely omitting geometric guidance on intermediate frames weakens the video model's inherent temporal continuity priors, thereby reducing the geometric stability of the final generated sequence.

\subsection{Limitation}
While our method demonstrates strong performance in camera-controllable image editing, two main limitations remain: (1) \textbf{Complex scenes and extreme viewpoint changes}  (in Fig.~\ref{fig:limition}) : When handling highly complex scenes or excessively large viewpoint variations, especially the latter, the introduced geometric references may become unreliable, leading to degraded geometric accuracy in the generated results; (2) \textbf{Inference efficiency}: Even with sparse temporal sampling to reduce the number of frames processed during inference, a certain number of frames still need to be generated. While this is significantly more efficient than standard video generation models, the inference time is still slightly longer compared to single-frame image diffusion models. Using a LoRA \cite{lightx2v} for accelerated sampling within the video model can further improve efficiency.

\section{Conclusion}
We propose \textbf{UniGeo}, a camera-controllable image editing framework that leverages the inherent continuity prior of video diffusion models to enforce unified geometric guidance throughout the generation process. By systematically integrating geometric guidance across representation, architecture, and loss function, UniGeo overcomes the limitations of fragmented geometric injection, establishing reliable cross-view correspondences while ensuring structural integrity. Comprehensive experiments demonstrate that UniGeo consistently outperforms existing methods in both geometric reliability and visual quality, providing a principled and effective solution for high-fidelity camera-controllable image editing across diverse camera motions.

\section*{Acknowledgements}
This work was supported in part by the Earth System Big Data Platform of the School of Earth Sciences, Zhejiang University.
\clearpage  


%
%
\bibliographystyle{splncs04}
\bibliography{main}

\clearpage

\end{document}